\documentclass[10pt,twocolumn,letterpaper]{article}

\usepackage{cvpr}              

\usepackage{graphicx}
\usepackage{amsmath}
\usepackage{amssymb}
\usepackage{booktabs}

\newcommand*\samethanks[1][\value{footnote}]{\footnotemark[#1]}

\usepackage[pagebackref,breaklinks,colorlinks]{hyperref}

\usepackage[capitalize]{cleveref}
\crefname{section}{Sec.}{Secs.}
\Crefname{section}{Section}{Sections}
\Crefname{table}{Table}{Tables}
\crefname{table}{Tab.}{Tabs.}

\def\cvprPaperID{*****} 
\def\confName{CVPR}
\def\confYear{2022}

\begin{document}

\title{Masked Autoencoders for Generic Event Boundary Detection

CVPR'2022 Kinetics-GEBD Challenge}

\author{\hspace{7mm} Rui He\thanks{Equal contribution.}\ \ \textsuperscript{1} \hspace{14mm} Yuanxi Sun\samethanks\ \ \textsuperscript{1} \hspace{10mm} Youzeng Li \samethanks\ \ \textsuperscript{1} \hspace{10mm} Zuwei Huang \samethanks\ \ \textsuperscript{1}\\
Feng Hu\textsuperscript{ 1}  \hspace{17mm} Xu Cheng\thanks{Corresponding Authors.}\ \ \textsuperscript{1 2} \hspace{14mm} Jie Tang\textsuperscript{\dag} \textsuperscript{2}\\
Tencent Holdings Ltd.\textsuperscript{1}\\
Department of Computer Science and Technology, Tsinghua University\textsuperscript{ 2}\\
{\tt\small \{rayruihe,yuanxisun,youzengli,takumihuang,emonhu,alexcheng\}@tencent.com}\\
\tt\small jietang@tsinghua.edu.cn}

\maketitle

\begin{abstract}
    Generic Event Boundary Detection (GEBD) tasks aim at detecting generic, taxonomy-free event boundaries that segment a whole video into chunks. In this paper, we apply Masked Autoencoders to improve algorithm performance on the GEBD tasks. Our approach mainly adopted the ensemble of Masked Autoencoders fine-tuned on the GEBD task as a self-supervised learner with other base models. Moreover, we also use a semi-supervised pseudo-label method to take full advantage of the abundant unlabeled Kinetics-400 data while training. In addition, we propose a soft-label method to partially balance the positive and negative samples and alleviate the problem of ambiguous labeling in this task. Lastly, a tricky segmentation alignment policy is implemented to refine boundaries predicted by our models to more accurate locations. With our approach, we achieved 85.94\% on the F1-score on the Kinetics-GEBD test set, which improved the F1-score by 2.31\% compared to the winner of the 2021 Kinetics-GEBD Challenge. Our code is available at \
    \href{https://github.com/ContentAndMaterialPortrait/MAE-GEBD}{\color{blue}{https://github.com/ContentAndMaterialPortrait/MAE-GEBD}}.
\end{abstract}

\section{Introduction}
\label{sec:intro}

\begin{figure*}
  \centering
  \includegraphics[width=\linewidth]{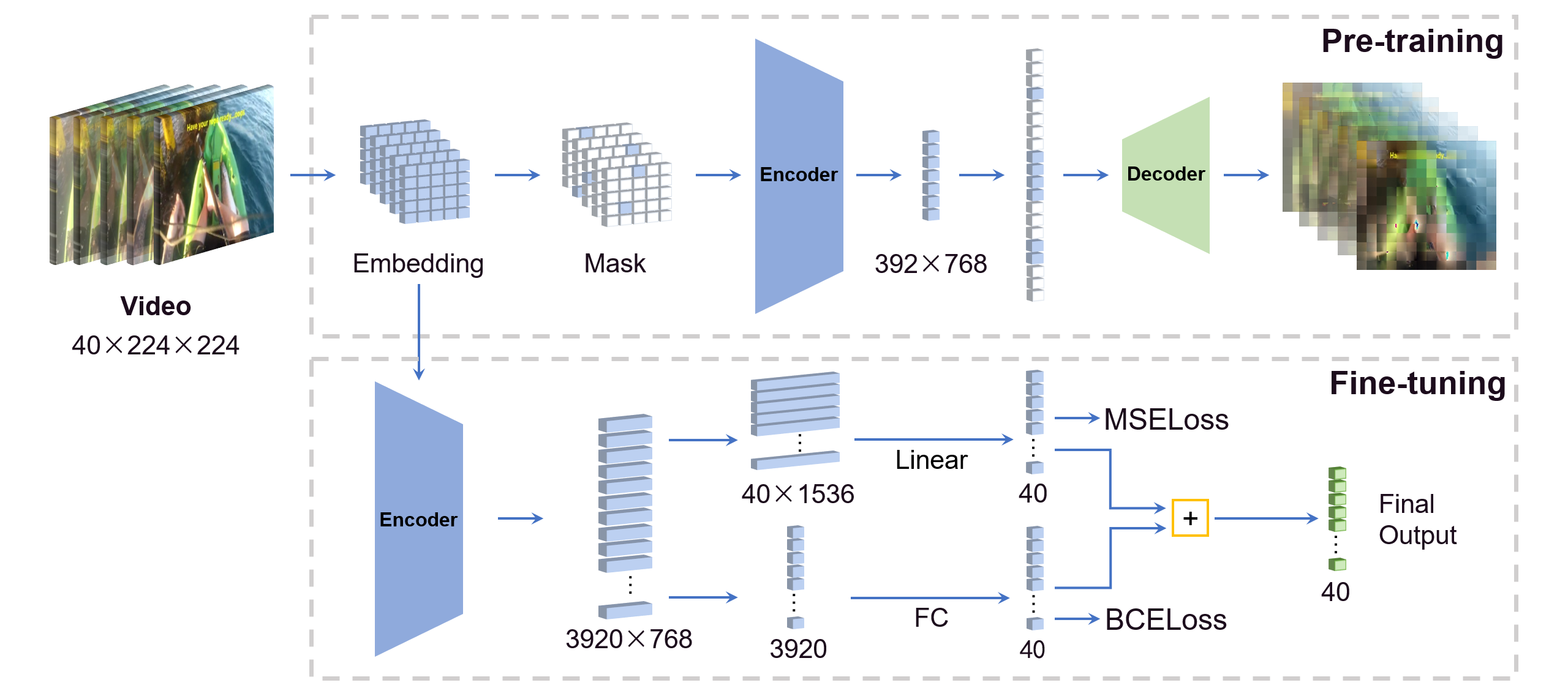}
  \caption{MAE-GEBD model structure. There are two stages in our MAE-GEBD model training. In the pre-training stage, we used the entire Kinetics-GEBD dataset for the pretraining to get embeddings, where 90\% of the data are masked out. In the second stage, we fine-tuned the encoder inherited from the previous stage with these embeddings without masks on the GEBD task. Here we deployed a multi-head training with both BCE and MSE losses, and the results from both are get averaged to get the final output. In particular, in the MSELoss branch, we apply a linear transformation on each 1536 feature to predict whether the corresponding position needs to be segmented.}
   \label{fig:model}
\end{figure*}

Vision is one of the most important ways for humans to obtain information. When people see continuous video images, they will first divide them into various segments according to the content of the video, so as to facilitate the understanding, paraphrasing, or other processes. Such mechanisms may be natural to the human brain, but not so easy for machine learning. In order for a model to mimic this information-gathering mechanism, the Generic Event Boundary Detection (GEBD) task was proposed to localize the boundaries automatically, around which the event content changes\cite{DBLP:journals/corr/abs-2101-10511}.

However, there are still many difficulties in the GEBD task. First, the definition of generic event boundaries is ambiguous compared to other computer vision tasks, such as image classification and object detection. This problem can be seen from the manual annotations in the Kinetics-GEBD datasets\cite{DBLP:journals/corr/abs-2101-10511} we used. Each video in Kinetics-GEBD has multiple different manual annotations, but the annotations from different annotators are quite different. The GEBD models, therefore, need to learn the commonality of various annotations. Moreover, the vague boundary definition also raises the difficulty and costs of the labeling process.

Currently, the mainstream models\cite{DBLP:journals/corr/abs-2101-10511,jwkim,DBLP:journals/corr/abs-2107-00239} solve the GEBD tasks by converting it into a binary classification task. The models cut the video into equal-length short segments. Each segment was treated as a positive sample if there was a boundary in the segment, and a negative sample otherwise. However, as there are about 5 boundaries labeled for every 40-segments video on average, the proportion between the positive and negative samples is around 1:7. This imbalance between positive and negative samples could make training more difficult. Moreover, when the model predicts that there is a boundary in a short segment, the boundary position will be locked at a fixed position according to the length of the segments.

Because of the difficulties mentioned above, we put forward  Pseudo-Label \cite{lee2013pseudo} to gain more training data without manual labeling. Meanwhile, we proposed soft label to alleviate the problem of ambiguous labeling and unbalanced positive and negative samples. And with a boundary alignment policy, the boundary position can be more accurately located. 

In addition to targeted solutions to existing difficulties, we also used MAE-based models to solve GEBD tasks from different perspectives. Our experiments show that using MAE-based GEBD models alone can already achieve comparable performance to the state-of-the-art. The performance based on existing models can be raised to a higher level with the help of the MAE-based models.

\section{Related Work}
\label{sec:related}

\subsection{Generic Event Boundary Detection}
Our model structure is largely based on \cite{jwkim}, who have won the Kinetics-GEBD challenge in 2021. They proposed the use of temporal self-similarity matrices (TSM) to solve the problem of event boundaries. The basic idea is that the video contents and hence their representations should vary significantly near the event boundaries, whereas these tend to remain unchanged away from the boundaries. In addition to the BinaryCrossEntropy loss, they also used contrastive learning as an auxiliary to learn better from the TSM results.

In general, this paper proposed a contrastive learning approach, a TSM approach, and a direct prediction approach without using TSM. The three branches are combined to form the entire model to make a final prediction. Our model structure is proposed mainly based on this model structure and modifications are made on this basis.

\subsection{Masked Autoencoders}
Self-supervised learners for computer vision has been heavily investigated in the literature\cite{he2020momentum, zbontar2021barlow,LocalMaskedReconstruction}. A model that is trained on broad data at scale and can be adapted (e.g., fine-tuned) to a wide range of downstream tasks\cite{FoundationModels} is called a foundation model. However, as data and models become larger and larger, the demands on the machine's performance also grow severely. How to efficiently and effectively train large models is one of the important topics of the foundation model research.

Recent Masked Autoencoders approaches have shown state-of-the-art performance on various tasks. MAE\cite{MAE} develop an asymmetric encoder-decoder architecture with masked tokens hiding a high proportion, e.g. 75\%, of the input images. The encoder operates only on the visible subset of image patches without masks and works with a lightweight decoder that reconstructs images. Multimodal masked autoencoder (M3AE\cite{MMAE}) consists of an encoder that maps language tokens and image patches to a shared representation space, and a decoder that reconstructs the original image and language from the representation. VideoMAE\cite{ VideoMAE} inspired by the ImageMAE and propose customized video tube masking and reconstruction, and show that video masked autoencoders are data-efficient learners for self-supervised video pre-training (SSVP). A conceptually simple extension of Masked Autoencoders (MAE) to spatiotemporal representation learning from videos suggests that the general framework of masked autoencoding (BERT, MAE, etc.) can be a unified methodology for representation learning with minimal domain knowledge\cite{MAESpatiotemporal}.

\section{Proposed Method}
\label{sec:method}

\subsection{Masked Autoencoders}
Following MAE, our Masked Autoencoder for Generic Event Boundary Detection(MAE-GEBD) is inspired by the recent ImageMAE\cite{MAE} and Spatiotemporal-MAE\cite{MAESpatiotemporal}.  In this challenge, we evenly take 40 frames for each 10s video and the video will be padded with its last frame if shorter than 10s. The video size is $40\times224\times224$ and the spatiotemporal patch size is $2\times16\times16$. In the pre-training stage, we refer to Spatiotemporal-MAE\cite{MAESpatiotemporal} and use a mask ratio of 90\%. Each sample has $20\times14\times14=3920$ tokens where 392 of which are visible. In the fine-tuning stage, we designed two headers for our model, using the BCELoss and MSELoss respectively with parameters inherited from the pre-training stage, as MSELoss can compensate for the potential insufficient gradient brought by BCELoss. Adding the MSELoss header yields a steady 0.4\% improvement in a single model of MAE-GEBD. \cref{fig:model} shows the MAE-GEBD model structure described above.

\subsection{Pseudo-Label}
Because the understanding of event boundaries is very subjective, there is a very huge gap between the definitions of boundaries among different annotators. The prediction results of the model trained by combining these annotations of different annotators could be even better than the individual manual annotations, so we thought of using the model to obtain more training data by labeling the unlabeled data using the trained models, and used these generated data to train the models even further. We obtain a variety of annotations for training by integrating different model ensemble methods and found that this Pseudo-Label strategy can improve model performance both on the validation and test set. 

Because of the limitation of time, we only annotate the model for the test set of the challenge. We believe that using more data from Kinetics-400 for pseudo-label training could help further improve the performance, which remains to be experimentally proven.

\subsection{Soft Label}

Similar to what has been done in  \cite{jwkim}, Our method divided each video into $p=0.25$ second sections, and the model could hence output a score for each section about whether we should segment or not. 

In \cite{jwkim}, each boundary annotation will only generate one positive sample with a target value of 1, and the other sections are negatives with 0. We believe that this will not only cause an imbalance of positive and negative samples but also make the task annotation becomes misleading, as we should still give credit if the model makes a prediction in the very neighborhood area of the annotation boundary. Therefore, we used soft label during the training process by allocating each annotation to the two sections nearby according to the ratio between the distance from the ground truth to each section. To be exact, if a marked boundary is located at the time $t$, the target scores should be:
$$\text{target}\left[\left\lfloor \frac{t}{p} \right\rfloor\right] = \left\lfloor \frac{t}{p} + 1\right\rfloor - \frac{t}{p} $$
$$\text{target}\left[\left\lfloor \frac{t}{p} \right\rfloor + 1\right] = \frac{t}{p} - \left\lfloor \frac{t}{p} \right\rfloor $$
where $\lfloor\ \cdot\ \rfloor$ is the round down computation.

After models output scores, we use a Gaussian filter to smoothen the score curve $s$, and use a max-pooling to output the final result. 
Because of the soft label method, the output score not only shows the possibility of a boundary at the current location but also reflects the distance between the exact boundary location and each output point location. Therefore, instead of directly outputting the local maximum of the score curve $s$, we will adjust the segment boundary by looking at the scores in a nearby window of ${\pm 2p}$ seconds. More specifically, the segment boundary will be shifted by a $bias$ computed as follows:
$$\text{bias}[i] = \left(\frac{\sum_{k=i+1}^{i+2}{s[k]} - \sum_{k=i-2}^{i-1}{s[k]}}{s[i]}\right) \times \frac{p}{2}$$
Generally speaking, the segment boundary will be adjusted according to the nearby two scores. The more unbalanced the scores are between each side, the closer the result is to the higher side.

\subsection{Segmentation Alignment}

\begin{figure}[t]
  \centering
   \includegraphics[width=\linewidth]{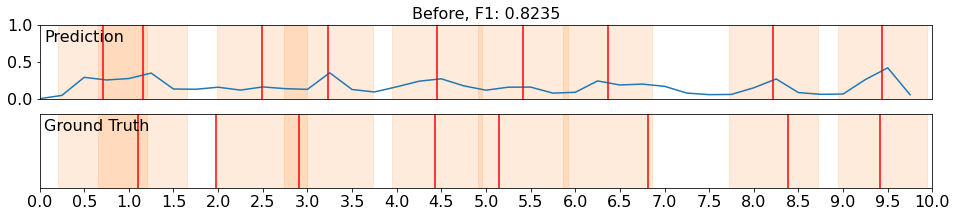}
   \includegraphics[width=\linewidth]{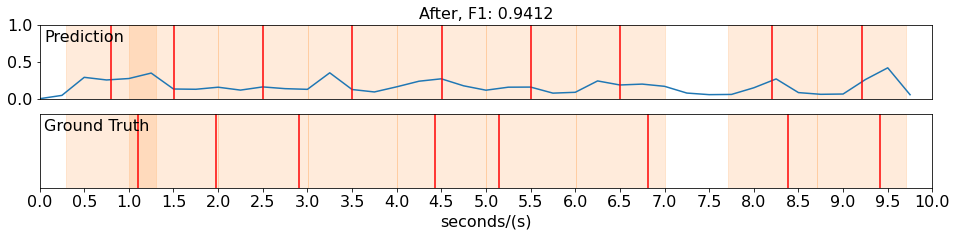}

   \caption{Example of the segmentation alignment policy, where the top-2 rows show the result and F1-score before the alignment, and the bottom-2 shows the performance after. The shaded area shows the $\pm$1s interval of each predicted boundary.}
   \label{fig:post}
\end{figure}

According to the definition of the challenge, there would not be a segment boundary in the first and last 0.3s of the video. And when computing the F1-score, a prediction will be counted as a true positive if there is a ground truth boundary that falls into a $[x\pm 5\% \times \text{duration}]$ interval around the prediction $x$. In order to make our predicted boundaries to be more efficient, we want the boundary intervals to have overlaps as small as possible. This means that our predictions should also be at least $[\pm 5\% \times \text{duration}]$ seconds away from each other, and away from the first and last 0.3s of the video as well. For example, if the video has a duration of 10s, then we will slightly shift the predicted boundaries so that $0.8\text{s} < x < 9.2\text{s}$ and $\|x_i - x_j\| > 1\text{s}$ for all prediction $x$. \cref{fig:post} shows one example of the segmentation alignment policy.

According to our experiment, 78.23\% of the videos will be affected by such segmentation alignment policy, and 9.17\% of which will get a 5\% improvement in terms of the F1-score, where only 2.28\% of which will get deducted by more than 5\%. Similar results appear no matter what model structures we used in the first place.

\section{Experiment}
\label{sec:experiment}

\subsection{Dataset and Feature}

We only used Kinetic-GEBD datasets for model training and validation. In the training process, because of the pseudo-label strategy, we not only used the train set (\textsuperscript{\texttildelow}17k) and valid set (\textsuperscript{\texttildelow}17k) of Kinetic-GEBD whose annotations coming from manual annotators, but also used the test set (\textsuperscript{\texttildelow}17k) of Kinetic-GEBD whose annotations are predicted by our various models. We mixed the train and valid data sets of Kinetics-GEBD, and then randomly divided them into 10 equal folds. One of them was selected as the validation set by turns, and the other nine as well as the Kinetics-GEBD test set predicted by the pseudo-label strategy were selected as the training set.

Our base models' input features are Kinetics pre-trained two-stream TSN features and SlowFast features, which are the same as \cite{jwkim}.

\subsection{Easy-Hard sample splits}
Hard example mining methods generally improve the performance of the object detectors, which suffer from imbalanced training sets\cite{HardExampleMining}. In this challenge, we found that the ratio of positive and negative samples in the dataset is also unbalanced. Therefore, we raise the positive weight parameter of the BCELoss to 8. Meanwhile, we assume that each video has at least one segment so that videos with a flat low-valued score curve ($s$) are regarded as poorly predicted hard samples. Therefore, we use the part of the datasets with the highest predicted score $\max(s)$ lower than threshold$+0.3$ (where threshold $=0.4$ in our method) as the training and validation test data for a Hard Model and others for an Easy Model. And we continue to train the model only on these subsets to make the model specialized in hard or easy cases. Our experiments show better results compared with other MAE-GEBD models.

\begin{table}
  \centering
  \begin{tabular}{lcc}
    \toprule
    \textbf{Method}        & \textbf{F1-Score} & \textbf{Improvement} \\
    \midrule
    Baseline               & 81.30             &                      \\
    soft label             & 81.80             & + 0.50               \\
    pseudo-label           & 82.10             & + 0.30               \\
    segmentation alignment & 83.20             & + 1.10               \\
    MAE-GEBD               & 83.06             &                      \\
    5-fold base ensemble   & 84.79             & + 1.59               \\
    10-fold base ensemble  & 84.89             & + 0.10               \\
    + MAE-GEBD             & 85.09             & + 0.20               \\
    + MAE-GEBD Easy        & 85.14             & + 0.05               \\
    \midrule
    \textbf{Test score}    & 85.94             &                      \\
    \bottomrule
  \end{tabular}
  \caption{Improvements of each method}
  \label{tab:example}
\end{table}

\subsection{Ensemble}
We tried to integrate our models using different approaches. However, we found that averaging the model output scores directly produced the best performance efficiently. So we tried different combinations of model structures and a different number of models in such an ensemble. Finally, we integrated 20 base models trained on 10 folds respectively and 4 MAE-GEBD models sharing the same data splits but using different training strategies.

More specifically, the 4 MAE-GEBD models consist of 2 models trained using the pseudo label method, 1 model without such strategy, and 1 model using the easy-hard sample splits methods. We split the test set into easy and hard samples using the criterion described above as well, apply the corresponding models,  and do the ensemble separately. According to our experiments, only the model fine-tuned on easy samples can further improve the results after the ensemble, so the easy subset goes through an ensemble of $20+4$ models, whereas the hard subset goes through only $20+3$.

\subsection{Experimental Results}
We improved our model performance gradually by adding those methods mentioned above. The improvements each method brings to the model on the validation set are shown in \cref{tab:example}. Eventually, we achieved 85.94\% on the F1-score on the Kinetics-GEBD test set.

As it is not possible for us to fully test the performance of the ensemble results with the validation set, some scores in this table are approximated from our test set submit performance. We just hope that different methods can be presented together to compare their improvements. It turns out that there is an inherent gap of about 0.8\% between the F1-score from the validation set and the test for our models, as we are always getting a higher test score in every submission. 

\section{Conclusion}
\label{sec:conclusion}

In this paper, we introduced our solution for the GEBD tasks. It not only made targeted strategies for the existing difficulties of GEBD, but also solved GEBD problems from multiple perspectives by integrating MAE models. Although our model has achieved very good performance, due to time constraints, there is still much we can do as future works to improve. For example, We could use more data to do MAE pre-training so as to give full play to its potential. We expected that the MAE-GEBD model could achieve better results than the basic models, but so far it does not get a higher F1-score and can only help after the ensemble. In addition, the pseudo-label strategy allows the model predictions to be generalized across the entire Kinetic-400 data set. We believe that adding more of the unlabeled data to the training could achieve better performance as well.

{\small
\bibliographystyle{ieee_fullname}
\bibliography{Paper}
}

\end{document}